\documentclass[11pt,a4paper]{article}
\usepackage[hyperref]{naaclhlt2019}
\usepackage{times}
\usepackage{latexsym}

\usepackage{url}
\usepackage{algorithm}
\usepackage{algorithmicx}
\usepackage[noend]{algpseudocode}

\aclfinalcopy 
\def\aclpaperid{1170} 

\usepackage{array}
\usepackage{amsmath}
\usepackage{amssymb}
\usepackage{amsfonts}

\usepackage{color}
\usepackage{multicol}
\usepackage{multirow}
\usepackage{balance}
\usepackage{url}
\usepackage{textcomp}

\usepackage{xspace}

\usepackage{helvet}

\usepackage{bm}
\usepackage{arydshln}

\usepackage{graphicx} 
\usepackage{graphics}
\usepackage{standalone}

\usepackage{caption}
\usepackage{subfig}

\usepackage{bbm}

\title{Information Aggregation for Multi-Head Attention with Routing-by-Agreement}

\author{Jian Li$^{1}$~~~~Baosong Yang$^2$~~~~Zi-Yi Dou$^3$~~~~Xing Wang$^4$~~~~Michael R. Lyu$^{1}$~~~~Zhaopeng Tu$^4$\thanks{~~~Zhaopeng Tu is the corresponding author of the paper. This work was mainly conducted when Jian Li, Baosong Yang, and Zi-Yi Dou were interning at Tencent AI Lab.}\\
\begin{tabular}{ccccc}
    \multicolumn{5}{c}{{\normalsize $^1$Department of Computer Science and Engineering, The Chinese University of Hong Kong}} \\
    \multicolumn{5}{c}{\em \normalsize \{jianli,lyu\}@cse.cuhk.edu.hk} \\
    {\normalsize $^2$University of Macau}   &   ~~~ &    {\normalsize $^3$Carnegie Mellon University}   &   ~~~ &   {\normalsize $^4$Tencent AI Lab} \\
    {\em \normalsize nlp2ct.baosong@gmail.com}  &   ~~~ &   {\em \normalsize zdou@andrew.cmu.edu}   &   ~~~ &   {\em \normalsize \{brightxwang,zptu\}@tencent.com} \\
\end{tabular}
  }

\begin{document}
\maketitle

\begin{abstract}
Multi-head attention is appealing for its ability to jointly extract different types of information from multiple representation subspaces. Concerning the information aggregation, a common practice is to use a concatenation followed by a linear transformation, which may not fully exploit the expressiveness of multi-head attention. In this work, we propose to improve the information aggregation for multi-head attention with a more powerful {\em routing-by-agreement} algorithm. Specifically, the routing algorithm iteratively updates the proportion of how much a part (i.e. the distinct information learned from a specific subspace) should be assigned to a whole (i.e. the final output representation), based on the agreement between parts and wholes. Experimental results on linguistic probing tasks and machine translation tasks prove the superiority of the advanced information aggregation over the standard linear transformation.
\end{abstract}

\section{Introduction}

Attention model becomes a standard component of the deep learning networks, contributing to impressive results in machine translation~\cite{Bahdanau:2015:ICLR,Luong:2015:EMNLP}, image captioning~\cite{xu2015show}, speech recognition~\cite{chorowski2015attention}, among many other applications. 
Its superiority lies in the ability of modeling the dependencies between representations without regard to their distance. Recently, the performance of attention is further improved by multi-head mechanism~\cite{Vaswani:2017:NIPS}, which parallelly performs attention functions on different representation subspaces of the input sequence. Consequently, different attention heads are able to capture distinct linguistic properties of the input, which are embedded in different subspaces~\cite{raganato2018analysis}. Subsequently, a linear transformation is generally employed to aggregate the partial representations extracted by different attention heads~\cite{Vaswani:2017:NIPS,Ahmed:2018:arXiv}.

Most existing work focus on extracting informative or distinct partial-representations from different subspaces~\citep[e.g.][]{lin2017structured,Li:2018:EMNLP}, while few studies have paid attention to the aggregation of the extracted partial-representations.
Arguably, information extraction and aggregation are both crucial for multi-head attention to generate an informative representation. Recent studies in multimodal learning show that a straightforward linear transformation for fusing features in different sets of representations usually limits the extent of abstraction~\cite{fukui2016multimodal,ben2017mutan}.
A natural question arises: {\em whether the straightforward linear transformation is expressive enough to fully capture the rich information distributed in the extracted partial-representations?}

In this work, we provide the first answer to this question. We propose to empirically validate the importance of information aggregation in multi-head attention, by comparing the performance of the standard linear function and advanced aggregation functions on various tasks. 
Specifically, we cast information aggregation as the {\em assigning-parts-to-wholes} problem~\cite{hinton2011transforming}, and investigate the effectiveness of the routing-by-agreement algorithm -- an appealing alternative to solving this problem~\cite{Sabour:2017:NIPS,Hinton:2018:ICLR}. 
The routing algorithm iteratively updates the proportion of how much a part should be assigned to a whole, based on the agreement between parts and wholes. We leverage the routing algorithm to aggregate the information distributed in the extracted partial-representations.

We evaluate the performance of the aggregated representations on both linguistic probing tasks as well as machine translation tasks. The probing tasks~\cite{conneau2018acl} consists of 10 classification problems to study what linguistic properties are captured by input representations. Probing analysis show that our approach indeed produces more informative representation, which embeds more syntactic and semantic information. 
For translation tasks, we validate our approach on top of the advanced \textsc{Transformer} model~\cite{Vaswani:2017:NIPS} on both WMT14 English$\Rightarrow$German and WMT17 Chinese$\Rightarrow$English data. Experimental results show that our approach consistently improves translation performance across languages while keeps the computational efficiency.

The main contributions of this paper can be summarized as follows:
\begin{itemize}
    \item To our best knowledge, this is the first work to demonstrate the necessity and effectiveness of advanced information aggregation for multi-head attention.
    \item Our work is among the few studies ({\it cf.} ~\cite{Gong:2018:arXiv,zhao2018investigating,Dou:2019:AAAI}) which prove that the idea of capsule networks can have promising applications on natural language processing tasks.
\end{itemize}

\section{Background}



Attention mechanism aims at modeling the relevance between representation pairs, thus a representation is allowed to build a direct relation with another representation.
Instead of performing a single attention function,~\newcite{Vaswani:2017:NIPS} found it is beneficial to capture different context features with multiple individual attention functions, namely multi-head attention.
Formally, attention function maps a sequence of query ${\bf Q}=\{{\bf q}_1,\dots, {\bf q}_J\}$ and a set of key-value pairs $\{{\bf K}, {\bf V}\}=\{({\bf k}_1, {\bf v}_1), \dots, ({\bf k}_M, {\bf v}_M)\}$ to outputs, where ${\bf Q} \in \mathbb{R}^{J \times d}$, $\{{\bf K}, {\bf V}\}\in \mathbb{R}^{M \times d}$. 
More specifically,  multi-head attention model first transforms $\bf Q$, $\bf K$, and $\bf V$ into $H$ subspaces with different, learnable linear projections:
\begin{equation}
  {\bf Q}_h, {\bf K}_h, {\bf V}_h   = {\bf Q}{\bf W}_h^{Q}, {\bf K}{\bf W}_h^{K}, {\bf V}{\bf W}_h^{V},
\end{equation}
where $\{{\bf Q}_h, {\bf K}_h, {\bf V}_h\}$ are respectively the query, key, and value representations of the $h$-th head. $\{{\bf W}_h^{Q}, {\bf W}_h^{K}, {\bf W}_h^{V}\} \in \mathbb{R}^{d \times \frac{d}{H}}$ denote parameter matrices associated with the $h$-th head, where $d$ represents the dimensionality of the model hidden states.
Furthermore, $H$ attention functions are applied in parallel to produce the output states $\{{\bf O}_1,\dots, {\bf O}_H\}$, among them:
\begin{equation}
  {\bf O}_h  = \textsc{Att}({\bf Q}_h, {\bf K}_h) {\bf V}_h,
\end{equation}
where ${\bf O}_h \in \mathbb{R}^{J \times \frac{d}{H}}$, $\textsc{Att}(\cdot)$ is an attention model. In this work, we use scaled dot-product attention~\cite{Luong:2015:EMNLP}, which achieves similar performance with its additive counterpart~\cite{Bahdanau:2015:ICLR} while is much faster and more space-efficient in practice~\cite{Vaswani:2017:NIPS}.


Finally, the $H$ output states are concatenated and linearly transformed to produce the final state:
 \begin{eqnarray}
 \text{Concat:} & \widehat{\bf O} =& [{\bf O}_1, \dots, {\bf O}_H]  , \label{eq:concat3}\\
  \text{Linear:} & {\bf O} =& \widehat{\bf O} {\bf W}^O,
  \label{eq:concat}
 \end{eqnarray}
where ${\bf O}\in \mathbb{R}^{J \times d}$ denotes the final output states, ${\bf W}^O \in \mathbb{R}^{d \times d}$ is a trainable matrix.

As shown in Equations \ref{eq:concat3} and \ref{eq:concat}, the conventional multi-head attention uses a straightforward concatenation and linear mapping to aggregate the output representations of multiple attention heads. 
We argue that this straightforward strategy may not fully exploit the expressiveness of multi-head attention, which can benefit from advanced information aggregation by exploiting the intrinsic relationship among the learned representations.


\section{Related Work}

Our work synthesizes two strands of research work, namely {\em multi-head attention} and {\em information aggregation}.

\subsection{Multi-Head Attention}

Multi-head attention has shown promising empirical results in many NLP tasks, such as machine translation~\cite{Vaswani:2017:NIPS,Domhan:2018:ACL}, semantic role labeling~\cite{Strubell:2018:EMNLP}, and subject-verb agreement task~\cite{Tang:2018:EMNLP}. The strength of multi-head attention lies in the rich expressiveness by using multiple attention functions in different representation subspaces. 

Previous work show that multi-head attention can be further enhanced by encouraging individual attention heads to extract distinct information. 
For example,~\newcite{lin2017structured} introduce a penalization term to reduce the redundancy of attention weights among different attention heads.~\newcite{Li:2018:EMNLP} propose disagreement regularizations to encourage different attention heads to capture distinct features, and~\newcite{Yang:2019:NAACL} model the interactions among attention heads.
~\newcite{Shen:2018:AAAI} explicitly use multiple attention heads to model different dependencies of the same word pair, and ~\newcite{Strubell:2018:EMNLP} employ different attention heads to capture different linguistic features.
Our approach is complementary to theirs, since they focus on extracting distinct information while ours aims at effectively aggregating the extracted information.
Our study shows that information aggregation is as important as information extraction for multi-head attention.

\subsection{Information Aggregation}

Information aggregation in multi-head attention (e.g. Equations~\ref{eq:concat3} and~\ref{eq:concat}) aims at composing the partial representations of the input captured by different attention heads to a final representation. Recent work shows that representation composition benefits greatly from advanced functions beyond simple concatenation or mean/max pooling. For example, \newcite{fukui2016multimodal} and \newcite{ben2017mutan} succeed on fusing multi-modal features (e.g., visual features and textual features) more effectively via employing the higher-order bilinear pooling instead of vector concatenation or element-wise operations. In NLP tasks,~\newcite{Peters:2018:NAACL} aggregate layer representations with linear combination, and~\newcite{Dou:2018:EMNLP} compose deep representations with layer aggregation and multi-layer attention mechanisms.

Recently, the routing-by-agreement algorithm, which origins from the capsule networks~\cite{hinton2011transforming}, becomes an appealing alternative to representation composition.
The majority of existing work on capsule networks has focused on computer vision tasks, such as MNIST tasks~\cite{Sabour:2017:NIPS,Hinton:2018:ICLR}, CIFAR tasks~\cite{xi2017capsule}, and object segmentation task~\cite{lalonde2018capsules}. 
The applications of capsule networks in NLP tasks, however, have not been widely investigated to date.~\newcite{zhao2018investigating} testify capsule networks on text classification tasks and~\newcite{Gong:2018:arXiv} propose to aggregate a sequence of vectors via dynamic routing for sequence encoding. 
~\newcite{Dou:2019:AAAI} use routing-by-agreement strategies to aggregate layer representations dynamically.
Inspired by these successes, we apply the routing algorithms to multi-head attention on both linguistic probing and machine translation tasks, 
which demonstrates the necessity and effectiveness of advanced information aggregation for multi-head attention.

\section{Approach}

\begin{figure}[t]
   \centering
  \includegraphics[width=0.48\textwidth]{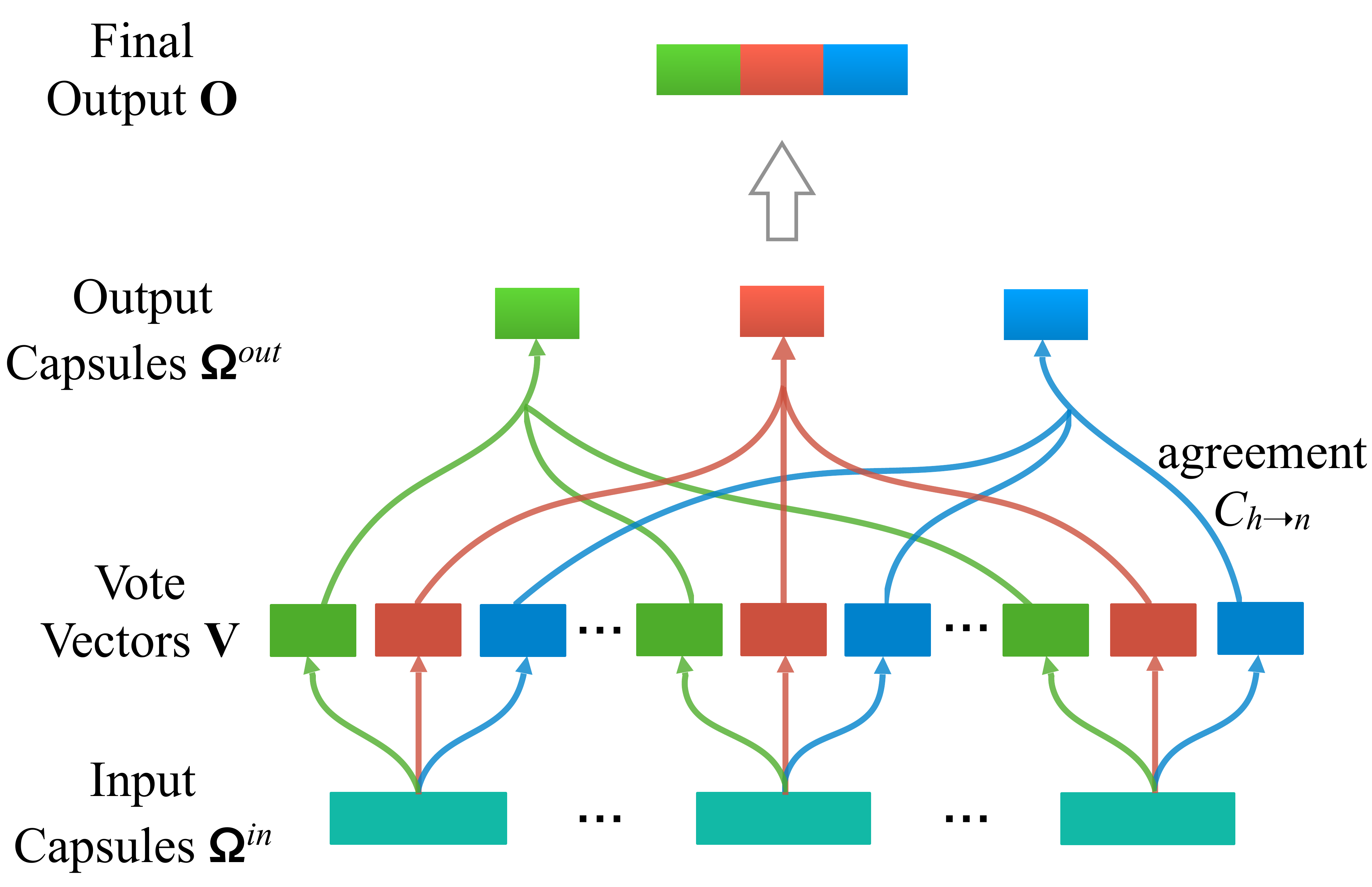}
 \caption{Illustration of routing-by-agreement.}
 \label{fig:routing}
\end{figure}

In this work, we cast information aggregation in multi-head attention as the problem of {\em assigning-parts-to-wholes}. Specifically, each attention head extracts different linguistic properties of the same input~\cite{raganato2018analysis}, and the goal of information aggregation is to compose the partial representations extracted by different heads to a whole representation. An appealing solution to this problem is the {\em routing-by-agreement} algorithm, as shown in Figure~\ref{fig:routing}.

The routing algorithm consists of two layers: {\em input capsules} and {\em output capsules}. The input capsules are constructed from the transformation of the partial representations extracted by different attention heads. 
For each output capsule, each input capsule proposes a distinct ``voting vector'', which represents the proportion of how much the information is transformed from this input capsule (i.e parts) to the corresponding output capsule (i.e. wholes). The proportion is iteratively updated based on the agreement between the voting vectors and the output capsule. Finally, all output capsules are concatenated to form the final representation.

\subsection{Routing-by-Agreement}

Mathematically, the input capsules ${\bf \Omega}^{in} = \{{\bf \Omega}^{in}_1, \dots, {\bf \Omega}^{in}_H\}$ with ${\bf \Omega}^{in} \in \mathbb{R}^{n \times d}$ are constructed from the outputs of multi-head attention:
\begin{equation}
    {\bf \Omega}^{in}_h = f_h (\widehat{\bf O}),
\end{equation}
where $f_h(\cdot)$ is a distinct non-linear transformation function associated with the input capsule ${\bf \Omega}^{in}_h$. Given $N$ output capsules, each input capsule ${\bf \Omega}^{in}_h$ propose $N$ ``vote vectors'' ${\bf V}_{h \rightarrow *} = \{{\bf V}_{h \rightarrow 1}, \dots, {\bf V}_{h \rightarrow N}\}$, which is calculated by
\begin{equation}
    {\bf V}_{h \rightarrow n} = {\bf \Omega}^{in}_h {\bf W}_{h \rightarrow n},
\end{equation}
Each output capsule ${\bf \Omega}^{out}_n$ is calculated as the normalization of its total input, which is a weighted sum over all ``vote vectors'' ${\bf V}_{* \rightarrow n}$:
\begin{equation}
    {\bf \Omega}^{out}_n = \frac{\sum_{h=1}^H C_{h \rightarrow n} {\bf V}_{h \rightarrow n}}{\sum_{h=1}^H C_{h \rightarrow n}}, \label{eqn:capsule}
\end{equation}
The weight $C_{h \rightarrow n}$ with $\sum_n C_{h \rightarrow n} = 1$ measures the agreement between vote vector ${\bf V}_{h \rightarrow n}$ and output capsule ${\bf \Omega}^{out}_n$, which is determined by the iterative routing as described in the next section. Note that $\sum_{h=1}^H C_{h \rightarrow n}$ is not necessarily equal to $1$. 
After the routing process, following~\newcite{Gong:2018:arXiv}, we concatenate the $N$ output capsules to form the final representation: ${\bf O} = [{\bf \Omega}^{out}_1, \dots, {\bf \Omega}^{out}_N]$. To make the dimensionality of the final output be consistent with that of hidden layer (i.e. $d$), we set the dimensionality of each output capsule be $\frac{d}{N}$.

\subsection{Routing Mechanisms}

In this work, we explore two representative routing mechanisms, namely {\em simple routing}~\cite{Sabour:2017:NIPS} and {\em EM routing}~\cite{Hinton:2018:ICLR}, which differ at how the agreement weights $C_{h \rightarrow n}$ are calculated.

\subsubsection{Simple Routing}

\begin{algorithm}[t]
\caption{\label{alg:simple} Iterative Simple Routing.}
\begin{algorithmic}[1]
\Procedure{Routing}{${\bf V}$, $T$}:
	\State{$\forall {\bf V}_{h \rightarrow *}$: $B_{h \rightarrow n} = 0$}
	\For{$T$ iterations}
		\State{$\forall {\bf V}_{h \rightarrow *}$:\ \ $C_{h \rightarrow n} =  \frac{\exp(B_{h \rightarrow n})}{\sum_{n'=1}^N\exp(B_{h \rightarrow n'})}$}
		\vspace{5pt}
		\State{$\forall \mathbf{\Omega}^{out}_n$:\ \ compute $\mathbf{\Omega}^{out}_{n}$ by Eq.~\ref{eqn:capsule}}
		\vspace{5pt}
		\State{$\forall {\bf V}_{h \rightarrow *}$:\ \ $B_{h \rightarrow n} \mathrel{+}= \mathbf{\Omega}^{out}_n \cdot {\bf V}_{h \rightarrow n}$}	
		\vspace{5pt}
	\EndFor
	\Return $\bf \Omega$
\EndProcedure
\end{algorithmic}
\end{algorithm}

Algorithm~\ref{alg:simple} lists a straightforward implementation of routing mechanism.
$B_{h \rightarrow n}$ measures the degree that the input capsule ${\bf \Omega}^{in}_h$ should be coupled to the output capsule ${\bf \Omega}^{out}_n$, which is initialized as all $0$ (Line 2). The agreement weights $C_{h\rightarrow n}$ are then iteratively refined by measuring the agreement between vote vector ${\bf V}_{h \rightarrow n}$ and output capsule ${\bf \Omega}^{out}_n$ (Lines 4-6), which is implemented as a simple scalar product $\mathbf{\Omega}^{out}_n \cdot {\bf V}_{h \rightarrow n}$ (Line 6).

To represent the probability that the output capsule ${\bf \Omega}^{out}_n$ is activated, \newcite{Sabour:2017:NIPS} use a non-linear ``squashing'' function: 
\begin{eqnarray}
    {\bf \Omega}^{out}_n &=& \frac{||{\bf \Omega}^{out}_n||^2}{1 + ||{\bf \Omega}^{out}_n||^2}  \frac{{\bf \Omega}^{out}_n}{||{\bf \Omega}^{out}_n||},     \label{eqn:squashing}
\end{eqnarray}

The scalar product $\mathbf{\Omega}^{out}_n \cdot {\bf V}_{h \rightarrow n}$ saturates at $1$, which makes it insensitive to the difference between a quite good agreement and a very good agreement. In response to this problem,~\newcite{Hinton:2018:ICLR} propose a novel Expectation-Maximization (EM) routing algorithm.

\subsubsection{EM Routing}

\begin{algorithm}[t]
\caption{\label{alg:em} Iterative EM Routing.}
\begin{algorithmic}[1]
\Procedure{EM Routing}{$\bf V$, $T$}:
    \State{$\forall {\bf V}_{h \rightarrow *}$: $C_{l \rightarrow n} = 1/N$}
	\For{$T$ iterations}
	    \State{$\forall \mathbf{\Omega}^{out}_n$:\ \ \textsc{M-step}(${\bf V}$, $C$)}   \Comment{{\small \em hold $C$ constant, adjust (${\bm \mu}_n, {\bm \sigma}_n, A_n$)}}
	    
	    \State{$\forall {\bf V}_{h \rightarrow *}$: \textsc{E-step}($\bf V$, ${\bm \mu}, {\bm \sigma}, A$)}   \Comment{{\small \em hold (${\bm \mu}, {\bm \sigma}, A$) constant, adjust $C_{h \rightarrow *}$}}
	\EndFor
	\State{$\forall \mathbf{\Omega}^{out}_n$: ${\bf \Omega}^{out}_n = A_n * {\bm \mu}_n$}
	
	\Return ${\bf \Omega}$
\EndProcedure
\end{algorithmic}
\end{algorithm}

Comparing with simple routing, EM routing has two modifications. First, it explicitly assigns an activation probability $A$ to represent the probability of whether each output capsule is activated, rather than the length of vector calculated by a squashing function (Equation~\ref{eqn:squashing}). Second, it casts the routing process as fitting a mixture of Gaussians using EM algorithm, where the output capsules play the role of Gaussians and the means of the input capsules play the role of the datapoints. Accordingly, EM routing can better estimate the agreement by allowing activated output capsules to receive a cluster of similar votes.

Algorithm~\ref{alg:em} lists the EM routing, which iteratively adjusts the means, variances, and activation probabilities (${\bm \mu}, {\bm \sigma}, A$) of the output capsules, as well as the agreement weights $C$ of the input capsules (Lines 4-5). The representation of output capsule ${\bf \Omega}^{out}_n$ is calculated as 
\begin{equation}
    {\bf \Omega}^{out}_n = A_n * {\bm \mu}_n = A_n * \frac{\sum_{h=1}^H C_{h \rightarrow n} {\bf V}_{h \rightarrow n}}{\sum_{h=1}^H C_{h \rightarrow n}},
\end{equation}

The EM algorithm alternates between an E-step and an M-step. The E-step determines, for each datapoint (i.e. input capsule), the probability of agreement (i.e. $C$) between it and each of the Gaussians (i.e. output capsules). The M-step holds the agreement weights constant, and for each Gaussian (i.e. output capsule) consists of finding the mean of these weighted datapoints (i.e. input capsules) and the variance about that mean.

\paragraph{\em M-Step} for each Gaussian (i.e. ${\bf \Omega}^{out}_n$) consists of finding the mean ${\bm \mu}_n$ of the votes from input capsules and the variance ${\bm \sigma}_n$ about that mean:
\begin{eqnarray}
    {\bm \mu}_n &=&\frac{\sum_{h=1}^H C_{h \rightarrow n} {\bf V}_{h \rightarrow n}}{\sum_{h=1}^H C_{h \rightarrow n}}, \label{eqn:mu} \\
    ({\bm \sigma}_n)^2 &=& \frac{\sum_{h=1}^H C_{h \rightarrow n} ({\bf V}_{h \rightarrow n}-{\bm \mu}_n)^2}{\sum_{h=1}^{H} C_{h \rightarrow n}}.  \label{eqn:sigma}
\end{eqnarray}
The incremental cost of using an active capsule ${\bf \Omega}^{out}_n$ is
\begin{eqnarray}
    \chi_n = \sum_{i} \big(\log({\bm \sigma}^i_n) + \frac{1+\log(2\pi)}{2}\big)\sum_{h=1}^H C_{h \rightarrow n}, \nonumber
\end{eqnarray}
where ${\bm \sigma}^i_n$ denotes the $i$-th dimension of the variance vector ${\bm \sigma}_n$.
The activation probability of capsule ${\bf \Omega}^{out}_n$ is calculated by 
\begin{equation}
    A_n = logistic\big(\lambda(\beta_A - \beta_{\mu} \sum_{h=1}^H C_{h \rightarrow n} - \chi_n)\big), \nonumber
\end{equation}
where $\beta_A$ is a fixed cost for coding the mean and variance of ${\bf \Omega}^{out}_n$ when activating it, $\beta_{\mu}$ is another fixed cost per input capsule when not activating it, and $\lambda$ is an inverse temperature parameter set with a fixed schedule. 
We refer the readers to~\cite{Hinton:2018:ICLR} for more details. 

%

\paragraph{\em E-Step} adjusts the assignment probabilities $C_{h \rightarrow *}$ for each input ${\bf \Omega}^{in}_h$. 
First, we compute the negative log probability density of the vote ${\bf V}_{h \rightarrow n}$ from ${\bf \Omega}^{in}_h$ under the Gaussian distribution fitted by the output capsule ${\bf \Omega}^{out}_n$ it gets assigned to:
\begin{equation}
    P_{h \rightarrow n} = \sum_i \frac{1}{\sqrt{ 2\pi({\bm \sigma}^i_n)^2}} \exp(- \frac{({\bf V}^i_{h \rightarrow n} - {\bm \mu}^i_n)^2}{2({\bm \sigma}^i_n)^2}). \nonumber
\end{equation}
Again, $i$ denotes the $i$-th dimension of the vectors $\{{\bf V}_{h \rightarrow n}, {\bm \mu}_n, {\bm \sigma}_n\}$.
Accordingly, the agreement weight is re-normalized by
\begin{equation}
    C_{h \rightarrow n} = \frac{A_n P_{h \rightarrow n}}{\sum_{n'=1}^N A_{n'} P_{h \rightarrow n'}}.
\end{equation}

\section{Experiments}
In this section, we evaluate the performance of our proposed models on both linguistic probing tasks and machine translation tasks.

\subsection{Linguistic Probing Tasks}

\begin{table*}[t]
  \centering
  \begin{tabular}{c|cc|ccc|ccccc}
     \multirow{2}{*}{\bf Model} & \multicolumn{2}{c|}{\bf Surface}  & \multicolumn{3}{c|}{\bf Syntactic}&\multicolumn{5}{c}{\bf Semantic}\\
    \cline{2-11}
        & SeLen & WC & TrDep & ToCo & BShif & Tense & SubNm & ObjNm & SOMO &CoIn \\
    \hline 
    \textsc{Base}  &\bf 97.22 & 97.92 & 44.48 & 84.44 & 49.30 & 84.20 & 87.66 & 82.94 & 50.24	& 68.77\\
    \hline
    \textsc{Simple}& 97.10 & \bf 98.85  &43.37   & 86.15  & 49.87  & \bf 88.22 & 87.25 &	85.07 & 48.77& 69.12\\
    \textsc{Em}& 96.26& 98.75&\bf 47.72&\bf \bf 87.00&\bf 51.82&88.17&\bf 89.97&\bf 86.40&\bf 51.55& \bf69.86   \\
  \end{tabular}
  \caption{Classification accuracies on 10 probing tasks of evaluating the linguistic properties (``Surface'', ``Syntectic'', and ``Semantic'') learned by sentence encoder. ``\textsc{Base}'' denotes the standard linear transformation, ``\textsc{Simple}'' is the simple routing algorithm, and ``\textsc{Em}'' is the EM routing algorithm.} 
  \label{tab:probing}
\end{table*}

\subsubsection{Setup}

\paragraph{Tasks}
Recently,~\newcite{conneau2018acl} designed 10 probing tasks to study what linguistic properties are captured by input representations.
A probing task is a classification problem that focuses on simple linguistic properties of sentences.
`SeLen' is to predict the length of sentences in terms of number of words. `WC' tests whether it is possible to recover information about the original words given its sentence embedding. `TrDep' checks whether an encoder infers the hierarchical structure of sentences. In `ToCo' task, sentences should be classified in terms of the sequence of top constituents immediately below the sentence node. `Bshif' tests whether two consecutive tokens within the sentence have been inverted. `Tense' asks for the tense of the main-clause verb. `SubNm' focuses on the number of the subject of the main clause. `ObjNm' tests for the number of the direct object of the main clause. In `SOMO', some sentences are modified by replacing a random noun or verb with another noun or verb and the classifier should tell whether a sentence has been modified. `CoIn' benchmark contains sentences made of two coordinate clauses. Half of the sentences are inverted the order of the clauses and the task is to tell whether a sentence is intact or modified. We conduct probing tasks to study whether the routing-based aggregation benefits multi-head attention to produce more informative representation.

\paragraph{Data and Models}
The models on each classification task are trained and examined using the open-source dataset provided by \newcite{conneau2018acl}, where each task is assigned 100k sentences for training and 10k sentences for validating and testing. 
Each of our probing model consists of 3 encoding layers followed by a MLP classifier. For each encoding layer, we employ a multi-head self-attention block and a feed-forward block as in \textsc{Transformer-Base}, which have achieved promising results on several NLP tasks~\cite{universal2018,bert2018}. The mean of the top encoding layer is served as the sentence representation passed to the classifier.
The difference between the compared models merely lies in the aggregation mechanism of multiple attention heads: ``\textsc{Base}'' uses a standard concatenation and linear transformation, ``\textsc{Simple}'' and ``\textsc{Em}'' are assigned simple routing and EM routing algorithms, respectively. For routing algorithms, the number of output capsules and routing iterations are empirically set to 512 and 3.

\subsubsection{Results}
Table~\ref{tab:probing} lists the classification accuracies of the three models on the 10 probing tasks. We highlight the best accuracies in bold.
Several observations can be made here. 

First, \emph{routing-based models produce more informative representation.} The representation produced by encoders with routing-based aggregation outperforms that by the baseline in most tasks, proving that routing mechanisms indeed aggregate attention heads more effectively. 
The only exception is the sentence length classification task (`SeLen'), which is consistent with the conclusion in \cite{conneau2018acl}: as a model captures deeper linguistic properties, it will tend to forget about this superficial feature.

Second, \emph{EM routing outperforms simple routing by embedding more syntactic and semantic information.} As shown in the last row, EM routing for multi-head aggregation consistently achieves best performances on most syntactic and semantic tasks. Especially on task `TrDep', `Tense' and `ObjNm',  EM routing-based model surpasses the baseline more than 3 points, demonstrating that EM routing benefits multi-head attention to capture more syntax structure and sentence meaning. 
Simple routing, however, underperforms the baseline model in some cases such as `TrDep' and `SubNm'. We attribute the superiority of EM routing to generating more accurate agreement weights with the Gaussian estimation.


\subsection{Machine Translation Tasks}

\begin{table*}[t]
  \centering
  \begin{tabular}{c||c|c|c|c||r|cc||c|c}
    \bf \#  &   \multicolumn{3}{c|}{\bf Applying Aggregation to \dots}   &   \multirow{2}{*}{\bf Routing}  &   \multirow{2}{*}{\bf \# Para.}     &   \multicolumn{2}{c||}{\bf Speed}   &   \multirow{2}{*}{\bf BLEU}  &  \multirow{2}{*}{$\bm \bigtriangleup$}\\  
    \cline{2-4}\cline{7-8}
    1   &   \em Enc-Self &   \em Enc-Dec   &   \em Dec-Self  &   &    &   Train   &   Decode  &   &\\
    \hline
    2   &   \texttimes  &   \texttimes   &  \texttimes  &  n/a    & 88.0M   &   1.92    &   1.67    &   27.31   &   --\\
    \hline
    3   &   \checkmark  &   \texttimes   &  \texttimes  & Simple  & +12.6M    &   1.23    &   1.66    &   27.98   &   +0.67\\
    \hdashline
    4   &   \checkmark  &   \texttimes   &  \texttimes  & EM      & +12.6M    &   1.20    &   1.65    &   28.28   &   +0.97\\
    5   &   \texttimes  &   \checkmark   &  \texttimes  & EM      & +12.6M    &   1.20    &     1.21      &   27.94   &   +0.63\\
    6   &   \texttimes  &   \texttimes   &  \checkmark  & EM      & +12.6M    &   1.21    &     1.21    &   28.15      & +0.84 \\
    \hdashline
    7   &   \checkmark  &   \checkmark   &  \texttimes  & EM      & +25.2M    &   0.87    &      1.20 &       28.45   &   +1.14\\
    8   &   \checkmark  &   \checkmark   &  \checkmark  & EM      & +37.8M    &   0.66    &   0.89    &   28.47   &   +1.16\\
  \end{tabular}
  \caption{Effect of information aggregation on different attention components, i.e., encoder self-attention (``{\em Enc-Self}''), encoder-decoder attention (``{\em Enc-Dec}''), and decoder self-attention (``{\em Dec-Self}''). ``\# Para.'' denotes the number of parameters, and ``Train'' and ``Decode'' respectively denote the training speed (steps/second) and decoding speed (sentences/second).}
  \label{tab:component}
\end{table*}

\begin{table}[t]
\centering
\begin{tabular}{c|c||rc|c}
\bf \# &  \bf Layers &  \bf \# Para.  &    \bf Train   &  \bf BLEU \\
\hline \hline
1  & None           &  88.0M    &   1.92  &   27.31\\
\hline
2  & \text{[1-6]}   &  100.6M   &   1.20  & 28.28 \\
\hdashline
3  & \text{[4-6]}   &  94.3M    &   1.54& 28.26\\
4  & \text{[1-3]}   &  94.3M    &   1.54& 28.27\\
\hdashline
5  &  \text{[1,2]}  &  92.2M    &   1.67  & 28.26\\
\hdashline
6  &  \text{[6]}    &  90.1M    &   1.88  & 27.68 \\
7  &  \text{[1]}    &  90.1M    &   1.88  & 27.75\\
\end{tabular}
\caption{Evaluation of different layers in the encoder, which are implemented as multi-head self-attention with the EM routing based information aggregation. ``1'' denotes the bottom layer, and ``6'' the top layer.}
\label{tab:layers1}
\end{table}

\begin{table*}[t]
  \centering
   \begin{tabular}{l|l||rl|rl}
    \multirow{2}{*}{\bf System}  &   \multirow{2}{*}{\bf Architecture}  &  \multicolumn{2}{c|}{\bf En$\Rightarrow$De}  & \multicolumn{2}{c}{\bf Zh$\Rightarrow$En}\\
    \cline{3-6}
        &   &   \# Para.    &   BLEU    &   \# Para.   &   BLEU\\
    \hline \hline
    \multicolumn{6}{c}{{\em Existing NMT systems}} \\
    \hline
    \cite{wu2016google} &   \textsc{GNMT}             &  n/a &  26.30   &   n/a &   n/a\\ 
    \cite{pmlr-v70-gehring17a}  &   \textsc{ConvS2S}  &  n/a &  26.36   &   n/a &   n/a\\
    \hline
    \multirow{2}{*}{\cite{Vaswani:2017:NIPS}} &   \textsc{Transformer-Base}    &  65M&   27.3  &   n/a   &   n/a \\ 
    &  \textsc{Transformer-Big}               &  213M   &  28.4 &   n/a&   n/a\\ 
    \hdashline
    \cite{hassan2018achieving}  &   \textsc{Transformer-Big}  &   n/a   &   n/a&   n/a&  24.2\\
    \hline\hline
    \multicolumn{6}{c}{{\em Our NMT systems}}   \\ \hline
    \multirow{4}{*}{\em this work}  &   \textsc{Transformer-Base}  &  88M&   27.31   &    108M&   24.13\\
    &   ~~ + Effective Aggregation   &  92M  &     28.26$^\Uparrow$   &  112M&  24.68$^\Uparrow$\\
    \cline{2-6}
    &   \textsc{Transformer-Big}     & 264M   &   28.58      &  304M   &   24.56\\
    &   ~~ + Effective Aggregation   & 297M   &   28.96$^\uparrow$  &  337M    &   25.00$^\uparrow$   \\
  \end{tabular}
  \caption{Comparing with existing NMT systems on WMT14 English$\Rightarrow$German (``En$\Rightarrow$De'') and WMT17 Chinese$\Rightarrow$English (``Zh$\Rightarrow$En'') tasks. 
  ``$\uparrow/\Uparrow$'': significantly better than the baseline counterpart ($p < 0.05/0.01$).}
  \label{tab:main}
  \end{table*}

\subsubsection{Setup}  
  
\paragraph{Data}
We conduct experiments on the widely-used WMT2014 English$\Rightarrow$German (En$\Rightarrow$De) and WMT2017 Chinese$\Rightarrow$English (Zh$\Rightarrow$En) machine translation tasks.
For the En$\Rightarrow$De task, the dataset consists of 4.6M sentence pairs. We use newstest2013 as the development set and newstest2014 as the test set. For the Zh$\Rightarrow$En task, we use all of the available parallel data, consisting of about 20.6M sentence pairs. We use newsdev2017 as the development set and newstest2017 as the test set.
We employ byte-pair encoding (BPE)~\cite{sennrich2016neural} with 32K merge operations for both language pairs.
We use the case-sensitive 4-gram NIST BLEU score~\cite{papineni2002bleu} as evaluation metric, and bootstrap resampling~\cite{Koehn2004Statistical} for statistical significance test.

\paragraph{Models}
We implement the proposed approaches on top of the advanced \textsc{Transformer} model~\cite{Vaswani:2017:NIPS}. 
We follow~\newcite{Vaswani:2017:NIPS} to set the configurations and have reproduced their reported results on the En$\Rightarrow$De task.
The \emph{Base} and \emph{Big} models differ at hidden size (512 vs. 1024) and number of attention heads (8 vs. 16). All the models are trained on eight NVIDIA P40 GPUs where each is allocated with a batch size of 4096 tokens.

\textsc{Transformer} consists of three attention components: encoder-side self-attention, decoder-side self-attention and encoder-decoder attention, all of which are implemented as multi-head attention. 
For the information aggregation in multi-head attention, we replace the standard linear transformation with the proposed routing mechanisms. We experimentally set the number of iterations to 3 and the number of output capsules as model hidden size, which outperform other configurations during our investigation.


\subsubsection{Component Analysis}

Table~\ref{tab:component} lists the results on the En$\Rightarrow$De translation task with \textsc{Transformer-Base}. As seen, the proposed routing mechanism outperforms the standard aggregation in all cases, demonstrating the necessity of advanced aggregation functions for multi-head attention.

\paragraph{Routing Mechanisms} (Rows 3-4)
We first apply simple routing and EM routing to encoder self-attention. Both  strategies perform better than the standard multi-head aggregation (Row 1), verifying the effectiveness of the non-linear aggregation mechanisms. Specifically, the two strategies require comparable parameters and  computational speed, but EM routing achieves better performance on translation qualities.
Considering the training speed and performance, {\em EM routing} is used as the default multi-head aggregation method in subsequent experiments.

\paragraph{Effect on Attention Components} (Rows 4-8)
Concerning the individual attention components (Rows 4-6), we found that the encoder and decoder self-attention benefit more from the routing-based information aggregation than the encoder-decoder attention. This is consistent with the finding in~\cite{Tang:2018:EMNLP}, which shows that self-attention is a strong semantic feature extractor.
Encouragingly, applying EM routing in the encoder (Row 4) significantly improve the translation quality with almost no decrease in decoding speed, which matches the requirement of online MT systems. We find that this is due to the auto-regressive generation schema, modifications on the decoder influence the decoding speed more than the encoder.

Compared with individual attention components, applying routing to multiple components (Rows 7-8) marginally improves translation performance, at the cost of a significant decrease of the training and decoding speeds. Possible reasons include that the added complexity makes the model harder to train, and the benefits enjoyed by different attention components are overlapping to some extent.
To balance translation performance and efficiency, we only apply EM routing to aggregate multi-head self-attention at the \emph{encoder} in subsequent experiments.

\paragraph{Encoder Layers}
As shown in Row 4 of Table~\ref{tab:component}, applying EM routing to all encoder layers significantly decreases the training speed by 37.5\%, which is not acceptable since \textsc{Transformer} is best known for both good performance and quick training. We expect applying to fewer layers can alleviate the training burden.
Recent studies show that different layers of NMT encoder can capture different levels of syntax and semantic features~\cite{shi2016emnlp,Peters:2018:NAACL}. Therefore, an investigation to study whether EM routing works for multi-head attention at different layers is highly desirable.

As shown in Table~\ref{tab:layers1}, we respectively employ EM routing for multi-head attention at the high-level three layers (Row 3) and low-level three layers (Row 4). The translation quality marginally drop while parameters are fewer and training speeds are quicker. This phenomena verifies that it is unnecessary to apply the proposed model to all layers. We further reduce the applied layers to low-level two (Row 5), the above phenomena still holds. However, a big drop on translation quality occurs when the number of layer is reduced to 1 (Rows 6-7).

Accordingly, to balance translation performance and efficiency, we only apply EM routing for multi-head aggregation at the \emph{low-level two layers of the encoder}, which we term ``Effective Aggregation'' in the following sections.

\subsubsection{Main Results}
In this section, we validate the proposed ``Effective Aggregation'' for multi-head attenion on both WMT17 Zh$\Rightarrow$En and WMT14 En$\Rightarrow$De translation tasks. The results are listed in Table~\ref{tab:main}. Our implementations of both \textsc{Transformer-Base} and \textsc{Transformer-Big} outperform the reported NMT systems on the same data and match the strong results of \textsc{Transformer} reported in previous works, which we believe make the evaluation convincing. Incorporating the effective aggregation consistently and significantly improves translation performance for both base and big \textsc{Transformer} models across language pairs, demonstrating the efficiency and universality of our proposed multi-head aggregation mechanism. 

Moreover, it is encouraging to see that  \textsc{Transformer-Base} with  effective aggregation strategy even achieves comparable performances to that of \textsc{Transformer-Big}, with about two thirds fewer parameters, which further demonstrates that our performance gains are not simply brought by additional parameters.

\section{Conclusion}

In this work, we provide first empirical validation on the importance of information aggregation for multi-head attention. Instead of the conventional linear transformation, we propose to aggregate the partial representations learned by multiple attention heads via {\em routing-by-agreement}. The routing algorithm iteratively updates the proportion of how much a partial representation should be assigned to the final output representation, based on the agreement between parts and wholes.

Experimental results across 10 linguistic probing tasks reveal that our EM routing-based model indeed produces more informative representation, which benefits multi-head attention to capture more syntactic and semantic information. In addition, our approach on various machine translation tasks consistently and significantly outperforms the strong \textsc{Transformer} baseline. Extensive analysis further suggests that only applying EM routing to low-level two layers of the encoder can best balance the translation performance and computational efficiency.

Future work includes combining our information aggregation techniques together with other advanced information extraction models for multi-head attention~\cite{Li:2018:EMNLP}.
We expect that the two kinds of approaches can complement each other to further improve the expressiveness of multi-head attention.

\section*{Acknowledgments}
Jian Li and Michael R. Lyu were supported by the Research Grants Council of the Hong Kong Special Administrative Region, China (No. CUHK 14210717 of the General Research Fund), and Microsoft Research Asia (2018 Microsoft Research Asia Collaborative Research Award).
We thank the anonymous reviewers for their insightful comments and suggestions.

\bibliography{all}
\bibliographystyle{acl_natbib}

\end{document}